\documentclass[letterpaper,twocolumn,10pt]{article}
\usepackage{usenix,epsfig,endnotes}
\begin{document}

\date{}

\title{\Large \bf Compilation as a Defense: Enhancing DL Model Attack Robustness via Tensor Optimization}

\author{
{\rm Stefan Trawicki}\\
Lancaster University
\and
{\rm William Hackett}\\
Lancaster University
\and
{\rm Lewis Birch}\\
Lancaster University
\and
{\rm Neeraj Suri}\\
Lancaster University
\and
{\rm Peter Garraghan}\\
Lancaster University, Mindgard
}

\maketitle

\thispagestyle{empty}

\subsection*{Abstract}
Adversarial Machine Learning (AML) is a rapidly growing field of security research, with an often overlooked area being model attacks through side-channels. Previous works show such attacks to be serious threats, though little progress has been made on efficient remediation strategies that avoid costly model re-engineering. This work demonstrates a new defense against AML side-channel attacks using model compilation techniques, namely tensor optimization. We show relative model attack effectiveness decreases of up to 43\% using tensor optimization, discuss the implications, and direction of future work.

\section{Introduction}

\subsection{Adversarial Machine Learning}
Adversarial Machine Learning (AML) investigates attacks on Deep Learning (DL) models and their underlying hardware/software assets
\cite{inversion, deepsniffer, zhu2020hermes, orekondy2018knockoff}. Of note are Side-Channel Accelerator (SCA) attacks which monitor kernel metrics during DL operation to extract leaky information \cite{deepsniffer, zhu2020hermes}. SCA attacks can be (1) model and dataset agnostic (black box), (2) conducted using few model inferences (< 1 second), and (3) notoriously complex to protect against. Current work remediates such attacks via costly model architecture and framework modifications \cite{modelobfuscate, deepsniffer}.

\subsection{Proposed Approach}

This work proposes a new defense against SCA attacks by utilizing generalized DL compilation optimization techniques to obfuscate model architecture, and increase model robustness to attacks. Popular DL frameworks such as PyTorch \cite{PyTorch} share common DL operator implementation libraries (e.g., CUDNN \cite{cudnn}). SCA attacks are designed to recognize model memory access patterns within these libraries, thus decreasing model architecture confidentiality. Our idea uses the TVM \cite{TVM} compiler to apply increasing levels of machine-generated tensor optimizations (TO) to obfuscate the model architecture without manual re-engineering. We demonstrate our defense approach effectiveness against a model SCA targeting kernel L2 cache on GPUs \cite{deepsniffer}.

AutoTVM is a simulated annealing technique for generating tensor optimization (TO) schedules, using a tuner to run $n$ candidate batches (trials) on the target hardware and guide the candidate search towards optimized schedules in the search space \cite{TVM}. More candidate trials typically result in faster-executing operators over time (e.g., Halide, Ansor \cite{halide, ansor}). We demonstrate AutoTVM using the XGB Rank cost-model \cite{xgbrankperformance} between 1 to 500 trials on YoloV4, RoBERTa, DenseNet121 and ResNet18 \cite{yolov4,roberta,densenet,resnet18}. These models range between 8m - 124m parameters, representing a diverse suite of applications (object detection, generative text, image classification). Robustness was assessed using DeepSniffer (DS) \cite{deepsniffer}, an SCA attack that associates kernel L2 cache reads/writes during inference to predict model architecture, selected due to having previously shown high reconstruction success \cite{hackett2023pinch}. ONNX models \cite{onnxmodels} were loaded into TVM, optimized, and inference performed in the TVM runtime with kernel metrics collected by the NSYS Profiler \cite{nsys}. DS was then run on the metrics to make architecture predictions with attack success measured as fidelity, the comparison of predicted and actual architecture between 0 (no similarity) and 1.0 (identical). Optimization and execution used CUDA 11.7 and the Nvidia A100 accelerator, with optimization code available on GitHub\footnote{https://github.com/stefanTrawicki/tdef}.

\section{Preliminary Findings}

\begin{figure}[h!]
  \centering
  \includegraphics[width=\linewidth]{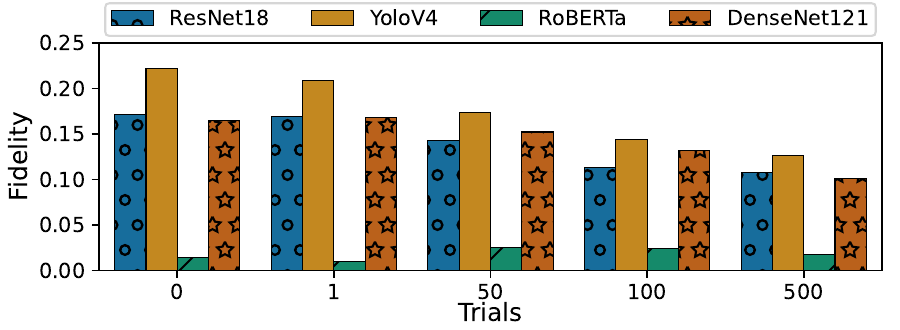}
  \caption{\textbf{DS fidelity over $n$ trials.} Compared to a 0 trial baseline, fidelity decreased with trials run.}
  \label{fig:trials_0}
\end{figure}

\noindent Initial findings demonstrate that compilation as a defense can successfully decrease the effectiveness of AML targeting the side-channel. Figure \ref{fig:trials_0} shows the impact of $n$ trials of TO on DS fidelity for each model. We observe that DS attack fidelity decreased with the number of trials conducted. After 500 trials, XGB decreased DenseNet121 fidelity from 0.1678 $\rightarrow$ 0.1013 (-39.6\%), and YoloV4 0.2220 $\rightarrow$ 0.1259 (-43.3\%). The failure of DS to effectively profile RoBERTa is attributed to it being a language model with heterogeneous operators unfamiliar to DS.

\section{Discussion} This work has investigated how tensor optimization techniques can reduce AML side-channel attack effectiveness by up to 43\%, without modifying model architecture. While maximizing optimization decreases inference time and improves robustness, it incurs large resource cost as generating and evaluating schedules is accelerator intensive (83 GPU-hours across all models/trials). In future work, we envision optimization for AML remediation via: (1) application with existing model-architecture modification approaches \cite{modelobfuscate}) to improve AML robustness, and (2) reactively applying optimization in response to AML detection, making the model a 'moving-target'. We also posit a focused technique where operators found to be conductive to SCA success are optimized first, decreasing resource requirements and attack effectiveness.

\newpage

{\footnotesize \bibliographystyle{acm}
\bibliography{sample}}

\end{document}